\documentclass[twocolumn]{article}

\usepackage{spconf,amsmath,graphicx,url, amssymb, booktabs, multirow}
\usepackage{makecell}
\usepackage{amsmath}
\usepackage{mdframed}
\usepackage[dvipsnames]{xcolor}
\usepackage{pdfpages}
\usepackage{graphicx}
\usepackage{array}
\usepackage{booktabs}
\usepackage{cite}
\usepackage{subcaption}
\usepackage{enumitem}

\title{ProGRes: Prompted Generative Rescoring on ASR N-Best}

\name{Ada Defne Tur$^{1,2}$, Adel Moumen$^{5}$, Mirco Ravanelli$^{2,3,4}$}
\address{$^{1}$McGill University,  $^{2}$Mila-Quebéc AI Institute, $^{3}$Concordia University, \\ $^{4}$Université de Montréal, $^{5}$Avignon Université\\ 
\texttt{ada.tur@mail.mcgill.ca}, \texttt{mirco.ravanelli@concordia.ca}, \\ \texttt{adel.moumen@univ-avignon.fr}}

\begin{document}

\maketitle
\begin{abstract}
Large Language Models (LLMs) have shown their ability to improve the performance of speech recognizers by effectively rescoring the $n$-best hypotheses generated during the beam search process.
However, the best way to exploit recent generative instruction-tuned LLMs for hypothesis rescoring is still unclear.
This paper proposes a novel method that uses instruction-tuned LLMs to dynamically expand the $n$-best speech recognition hypotheses with new hypotheses generated through appropriately-prompted LLMs. Specifically, we introduce a new zero-shot method for ASR $n$-best rescoring, which combines confidence scores, LLM sequence scoring, and prompt-based hypothesis generation.
We compare Llama-3-Instruct, GPT-3.5 Turbo, and GPT-4 Turbo as prompt-based generators with Llama-3 as sequence scorer LLM. 
We evaluated our approach using different speech recognizers and observed significant relative improvement in the word error rate (WER) ranging from 5\% to 25\%.
\end{abstract}
\vspace*{-2ex}
\section{Introduction}
Despite the significant advancements of recent years, Automatic Speech Recognition (ASR) still poses some challenges \cite{yu2018recent}. These challenges are particularly evident in scenarios involving noise and reverberation \cite{li2014overview, kinoshita2016summary,noise_cont,reverb,matassoni2014dirha,ravanelli2017deep}, spontaneous dialogues \cite{cornell2023chime7,info14020137, mousavi2024are}, or recordings containing numerous named entities such as people, locations, organizations, and scientific/technical terminology \cite{liu2011entity, caubriere-etal-2020-named, mdhaffar2022end}.
ASRs are indeed not trained on enough data to deeply capture linguistic information, resulting in challenges when transcribing unknown words and named entities, particularly those related to social culture (such as ``Billie Eilish" or ``Tour de Ski").

\begin{figure}[t!]
\begin{footnotesize}
    \noindent \textbf{ASR $n$-best:}
    \begin{itemize}[leftmargin=*]
        \itemsep -0.5ex
        \item it is distributed throughout the southern philippines {\color{red}walaca} and new {\color{red}guina}
        \item it is distributed throughout the southern {\color{red}pilippines walaca} and new {\color{red}guina}
        \item it is distributed throughout the southern philippines {\color{red}valaca} and new {\color{red}guina}
        \item it is distributed throughout the southern {\color{red}pilippines valaca} and new {\color{red}guina}
        \item it is distributed throughout the southern philippines {\color{red}walaka} and new {\color{red}guina}
        \end{itemize}
    \noindent \textbf{Prompt Addition:} {it is distributed throughout the southern {\color{ForestGreen}philippines wallacea} and new {\color{ForestGreen}guinea}\\
    \noindent \textbf{Ground Truth:} it is distributed throughout the southern {\color{ForestGreen}philippines wallacea} and new {\color{ForestGreen}guinea}}
\end{footnotesize}
\vspace*{-2ex}
\caption{Example of a prompted generation using ASR $n$-best.}
\label{fig:ex}
\vspace*{-1ex}
\end{figure}

To mitigate these issues, language models are commonly used to enhance ASR performance by ensuring that transcriptions maintain linguistic plausibility.
The dominant approach for combining acoustic and linguistic information is known as \textit{rescoring}, which involves using a language model to assign probabilities to candidate word sequences \cite{asr-n-best-rescoring, languages7030236, 9747118, udagawa2022effect, 9679032, Xu2018APR, chung-etal-2012-lattice, 6707767, DBLP:journals/corr/abs-1210-8440}. These probabilities are then combined linearly with the acoustic scores provided by the ASR. Rescoring can be performed in two ways: either during the decoding process, where partial hypotheses are rescored, or after decoding, where the $n$-best alternatives are rescored \cite{Sak2010OntheflyLR, ma23}.
Traditionally, rescoring involved using simple language models, such as n-grams and word graphs \cite{ORTMANNS199743, hakkani2003}. Only relatively recently, recurrent neural networks and transformers have been successfully employed \cite{vaswani2023attention}.
Modern LLMs, on the other hand, have achieved remarkable success in natural language processing, making them ideal candidates for ASR rescoring. LLMs have much vaster knowledge of a human language than an ASR and can predict how phonetic patterns between arbitrary sets of words can correlate to more meaningful replacements. 
Consider the example depicted in Figure~\ref{fig:ex}: only a powerful LLM has the information necessary to identify relevant locations and correct the hypotheses.

\begin{figure*}[t!]
\centering
\includegraphics[width=0.8\textwidth]{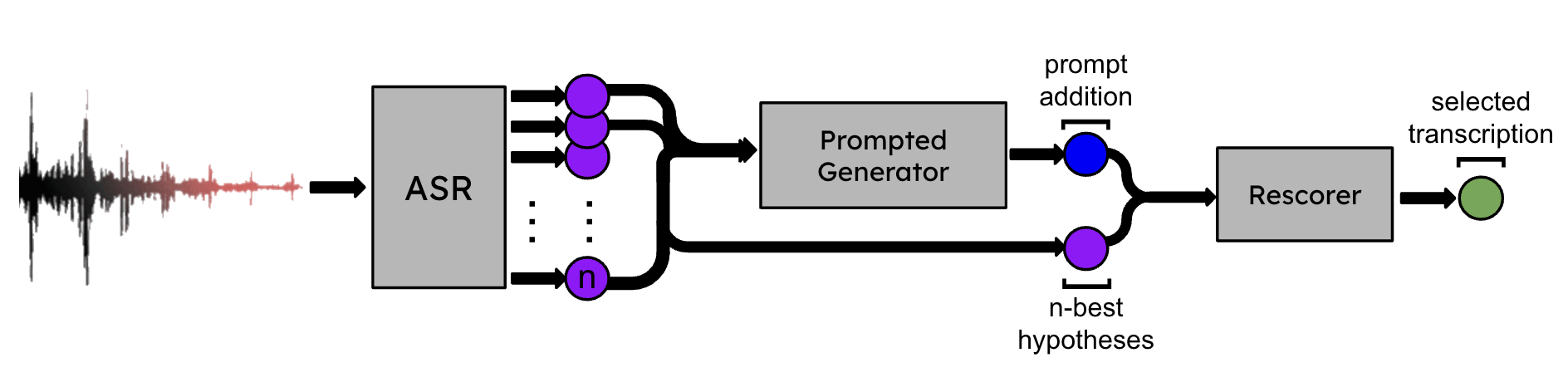}
\vspace*{-2ex}
\caption{Overview of the ProGRes pipeline: (1) An ASR generates $n$-best hypotheses. (2) The hypotheses are extended with a suggestion from an LLM. (3) The entire set is rescored to produce the final transcription.}
\label{fig:pipeline}
\vspace*{-2ex}
\end{figure*}

The optimal way to leverage instruction-tuned LLMs for ASR rescoring is the object of intense research efforts.
In \cite{ma2023generative}, for instance, the authors proposed prompting an LLM like Chat-GPT to select from the $n$-best options or generate a new output, demonstrating significant effectiveness in improving ASR error rates.

This paper extends this approach by introducing a novel method to effectively use LLMs for ASR $n$-best rescoring, called  PROmpted Generative REScoring (ProGRes). In particular, our main contributions are the following:
\begin{itemize}[leftmargin=*]
\itemsep -0.5ex
    \item We propose using prompt-based LLMs to dynamically augment the ASR $n$-best, informed by its existing hypotheses. Our idea originated from observing that rescoring provides minimal benefits, even with LLMs, when the correct transcription is not among the top $n$ hypotheses. With prompt addition, we aim to increase the likelihood that at least one hypothesis closely matches the correct transcription, reducing the overall WER significantly.

    \item Unlike previous work which ignores the existing ASR $n$-best hypotheses after generating the LLM ones~\cite{ma2023generative}, we propose to assign a score to each hypothesis in the extended set using another state-of-the-art open-weight LLM (in our case, Llama-3). These LLM probabilities are then interpolated with the ASR confidence scores.
\end{itemize}

We demonstrate that ProGRes leads to significant improvements in word error rates. It also outperforms existing methods that restrict the use of generative LLMs to either $n$-best rescoring or generating a new hypothesis alone. 

The code for implementing ProGRes and replicating the experiments discussed in this paper is publicly available on GitHub\footnote{\url{https://github.com/AdaDTur/ProGRes/}}.

\vspace*{-2ex}
\section{Proposed Method}
The proposed method is summarized in Figure \ref{fig:pipeline}. First, for each audio signal, we extract the $n$-best hypotheses from a pretrained ASR model. We then prompt an LLM to generate a more diverse set of hypotheses. Subsequently, we rescore these extended hypotheses to obtain the final transcription.
The following sub-section will provide a detailed description of the proposed method.
\vspace*{-1ex}
\subsection{Prompted Generation}
The $n$-best hypotheses generated by the ASR are often noisy and sub-optimal. Many of these hypotheses rely solely on phonetic evidence and do not incorporate general knowledge, common sense, or deep linguistic insights.
Our approach aims to improve the quality of the $n$-best hypotheses by using a secondary generative LLM. This intelligent model analyzes the $n$-best hypotheses and generates a more plausible transcription, that avoids non-existent tokens, words, named entities, etc.
Similar to \cite{ma2023generative}, we use the following prompt:

\scriptsize
\begin{verbatim}
"Perform error correction on the top outputs generated 
by an ASR system. The ASR hypotheses, listed in order of
their ASR posterior score, are as follows: [ASR n-best] 
Please provide the corrected ASR transcription of the 
given utterance only, surrounding it with < >. Do not 
add any explanations or commentary."
\end{verbatim}
\normalsize

For each set of hypotheses, the LLM is prompted to predict the most plausible transcription. The LLM has the flexibility to either select an existing hypothesis or generate one entirely from scratch. The ASR $n$-best hypotheses and the LLM-generated hypothesis are concatenated to form an extended set of hypotheses.
\vspace*{-2ex}
\subsection{LLM score}
The next step involves assigning a linguistic plausibility score to each hypothesis within the extended set. This score can be computed by employing an open-weight LLM, such as Llama-3, and using the following equation:
\[
LLM_{score}(\mathbf{w}) = \sum_{t=1}^{T} \log P_{\text{LLM}}(w_t | w_1, w_2, \ldots, w_{t-1}; \boldsymbol{\theta}).
\]
Here \( LLM_{score} \) denotes the pseudo-log-likelihood score for the sequence \( \mathbf{w} = [w_1, w_2, \ldots, w_t, \ldots, w_T]^T \), where \( w_t \) is the token at position \( t \). The term \( P_{\text{LLM}} \) is the conditional probability of token \( w_t \) given the preceding tokens, while \( \boldsymbol{\theta} \) are the parameters of the considered LLM. A higher \( LLM_{score} \) indicates greater linguistic plausibility of the hypothesis
\vspace*{-2ex}
\subsection{ASR score}
\label{sec:ASR_score}
For each hypothesis in the extended set, we also compute the ASR confidence score by evaluating the loss of the ASR model with the hypothesis treated as the target label:
\[ ASR_{score}(\mathbf{w}, \mathbf{x}) = - \text{loss}(p(\mathbf{x}), \mathbf{w}), \]
where \( p(\mathbf{x}) \) represents the sequence of token probabilities predicted by the ASR model with the audio signal $\mathbf{x}$ in input and $\mathbf{w}$ is the hypothesis for which we want to determine the score. 
The ASR score depends on ``how well" the hypothesis matches the acoustic evidence. The $loss$, which should be in the log-domain, can either be the Negative Log-Likelihood (NLL) or the Connectionist Temporal Classification (CTC) loss, or any other loss used to train the ASR model. Note that during training, the loss is normally minimized. Therefore, we need to negate it to get a score that is higher when the hypothesis matches the acoustic signal well.

The ASR score can be computed for all hypotheses, including those generated by the LLM. We assign the confidence score of the best ASR hypothesis to the LLM-generated one. The LLM-generated hypothesis is more likely to be the best, and we observed that boosting it in this way helps improve performance.
\vspace*{-2ex}
\subsection{Interpolation}
\label{sec:interpolation}
To combine LLM and ASR scores, we use linear interpolation, as shown in the following equation:
\[
\text{score} = (1 - \alpha) \cdot \text{ASR}_{\text{score}} + \alpha \cdot \text{LLM}_{\text{score}},
\]
where the hyperparameter \(\alpha\) is introduced to adjust the ratio between the ASR and LLM scores. The optimal value of the language model weight \(\alpha\) can be determined by performing a linear search optimization on the validation set. This step might involve testing values between 0 and 1 in increments of 0.05, as discussed in Sec.~\ref{sec:int}. The hypothesis with the highest score is selected as the output ASR transcription.

\vspace*{-2ex}
\section{Related Work}
Masked language models (MLMs), with their popular speech-specific variant, warped language models (WLMs), have been adopted to improve ASR performance. MLMs can generate embeddings that, when combined with acoustic and linguistic information, serve as input for the rescorer. Additionally, MLMs can be employed to extract sequence scores \cite{fohr21}. WLMs can analyze an entire sequence of all $n$-best hypotheses and perform token deletion, insertion, and substitution as needed to rescore the $n$-best hypotheses provided by the ASR \cite{namazifar2021correcting}.
Ma et al. explored the use of generative models for rescoring, particularly architectures like T5 \cite{ma23}. Instead of using a masked model to analyze and alter the hypotheses, their approach involves using a fine-tuned rescoring model fed by either the $n$-best hypotheses or the ASR beam-search decoding lattice.

While fine-tuned rescoring models have proven effective in enhancing ASR performance, the emergence of prompt-based models recently introduced new opportunities for ASR error correction.
Building on their previous work, Ma et al. investigated the potential of prompt-based language models \cite{ma2023generative}. They employed two methods: a selective approach where the model, such as ChatGPT, analyzes the $n$-best hypotheses and selects the most suitable option, and an unconstrained approach where ChatGPT's generative capabilities are exploited to correct possible errors.
Our contribution extends the work by Ma et al. in different ways. First, ProGRes concatenates the $n$-best ASR hypotheses and those generated by the LLMs. This step ensures that the considered hypotheses does not diverge significantly from the original ASR ones.
Additionally, unlike their approach, we incorporate a rescoring step that interpolates ASR and LLM scores. These improvements have resulted in superior ASR performance.

\vspace*{-2ex}
\section{Experimental Setup}
\vspace*{-2ex}
\subsection{ASRs}
Our study considers two speech recognizers. First, we employ an ASR trained on SPGISpeech data using the ESPnet toolkits \cite{watanabe2018espnet, hayashi2020espnet} (ASR$_1$). This recognizer utilizes an STFT with 512 FFT points and a hop length of 256 to extract spectrogram features. It employs a conformer architecture as an encoder, which combines convolutional front-end for local\ dependencies with a self-attention mechanism for longer-range dependencies. 
Additionally, we utilize an ASR system trained on CommonVoice 16.1 using SpeechBrain \cite{ravanelli2021speechbrain}\footnote{\url{https://github.com/speechbrain/speechbrain/tree/develop/recipes/CommonVoice/ASR}} (ASR$_2$). This recognizer is constructed using WavLM for feature extraction, a standard deep neural network, and connectionist temporal classification (CTC) for sequence modeling \cite{baevski2020wav2vec}. 
The decoding process for both ASRs involves beam search to generate a set of $n$ hypotheses with associated probabilities \cite{gulati2020conformer, sennrich-etal-2016-neural}.
We set n=10 for ASR$_1$ and n=20 for ASR$2$.
\vspace*{-2ex}
\subsection{LLMs}
We evaluate several LLMs for generating and rescoring prompted hypotheses. Specifically, for prompting, we assess three instruction-tuned models: GPT-3.5-turbo, GPT-4 \cite{openai2024gpt4technicalreport}, and Llama-3 \cite{brown2020language}. We utilize the OpenAI API for optimized generation with the GPT models. For Llama-3, we employ the quantized version provided by UnslothAI\footnote{\url{https://www.unsloth.ai/blog/llama3}}.
For rescoring, we require an open-weight model that allows us to compute the LLM log probabilities. We utilize the standard non-instruction-tuned Llama-3 model from Meta, which has 8 billion parameters \cite{touvron2023llama}. Importantly, we have not fine-tuned any models throughout our process, resulting in a purely zero-shot pipeline.

\begin{table*}[t!]
    \centering
    \begin{tabular}{lcccc}
    \toprule
     & \textbf{Prompt Only} & \textbf{LLM Rescoring} & \textbf{ProGRes} & \textbf{Oracle} \\
     & ASR$_1$ $\mathbf{|}$ ASR$_2$ & ASR$_1$ $\mathbf{|}$ ASR$_2$ & ASR$_1$ $\mathbf{|}$ ASR$_2$ & ASR$_1$ $\mathbf{|}$ ASR$_2$ \\
    \midrule
    \textbf{ASR $n$-best Hypotheses} & - & 43.75 $\mathbf{|}$ 13.97 & 42.84 $\mathbf{|}$ 11.42 & 38.38 $\mathbf{|}$ 6.82 \\
    \midrule
    \textbf{ASR + Llama 3 Prompt} & 49.12 $\mathbf{|}$ 11.41 & 46.73 $\mathbf{|}$ 13.61 & 46.14 $\mathbf{|}$ 11.30 & 37.42 $\mathbf{|}$ 6.54 \\
    \midrule
    \textbf{ASR + GPT 3.5 Prompt} & 46.05 $\mathbf{|}$ 10.50 & 44.82 $\mathbf{|}$ 12.98 & 44.19 $\mathbf{|}$ 10.57 & 37.18 $\mathbf{|}$ 6.62 \\
    \midrule
    \textbf{ASR + GPT 4 Prompt} & 42.65 $\mathbf{|}$ 9.39 & 41.70 $\mathbf{|}$ 11.80 & \textbf{40.84 $\mathbf{|}$ 9.32} & 36.17 $\mathbf{|}$ 5.48 \\
    \bottomrule
    \end{tabular}
    \caption{WER on the CommonVoice test set using various combinations of prompting, rescoring, and ASR models. All values are percentages.}
    \label{table:1}
\end{table*}
\vspace*{-2ex}
\subsection{Evaluation Dataset}
The performance of ASR$_1$ is evaluated using the test set of CommonVoice 14.0 (English portion), that of ASR$_2$ on the test set of CommonVoice 16.1 (English portion), both comprising ~16,300 audio samples \cite{commonvoice:2020}. It is important to note that the ASR$_1$ operates under mismatch conditions as it is trained on data different from CommonVoice, whereas the ASR$_2$ is trained specifically on CommonVoice data. This setup enables us to explore various experimental conditions.
The hyperparameter \(\alpha\) is tuned using the validation set.

\vspace*{-2ex}
\subsection{Libraries}
Beyond the SpeechBrain \cite{ravanelli2021speechbrain} and ESPnet toolkit \cite{watanabe2018espnet}, we used TorchAudio for audio file reading \cite{torchaudio21}. We use the Minicons library to extract the LLM score for each hypothesis \cite{misra2022minicons}, without altering the casing or punctuation. We use the OpenAI python API for GPT-3.5 and GPT-4 models\footnote{\url{https://github.com/openai/openai-python}}.

\vspace*{-2ex}
\section{Results}
This section shows the experimental validation of the proposed ProGRes method for rescoring.
\vspace*{-2ex}
\subsection{Baselines} 
We compute the baseline WER for both speech recognizers under consideration. For ASR$_1$, we achieve a WER of 42.94\% on the test set. This high WER is due to the mismatched conditions, as the model is trained on SPGISpeech (which contains audio data from the financial domain) and tested on CommonVoice (which includes data from various other domains).
For ASR$_2$, the WER is 12.38\% on the test set. In this case, the model is both trained and tested on CommonVoice data, using the training and test portions of the dataset, respectively.
\vspace*{-2ex}
\subsection{Prompted Generative Rescoring}
\label{sec:prompted}
Table \ref{table:1} presents the ASR performance of both ASR$_1$ and ASR$_2$ on the test set under various configurations: 

\begin{itemize} 
\itemsep -0.0ex
\item \textbf{Prompt Only}: This configuration uses the LLM-generated hypothesis as the ASR transcription, similar to the unconditioned generation approach discussed in Ma et al. 
\item \textbf{LLM Rescoring}: Here, an extended set of hypotheses, including both ASR $n$-best and LLM-generated ones, is reranked based on LLM scores, with the top-scoring hypothesis selected as the ASR transcription. 
\item \textbf{ProGRes}: This is the proposed method, which reranks the extended set of hypotheses using an interpolation between LLM and ASR scores. 
\item \textbf{Oracle}: This configuration selects the hypothesis with the lowest WER from the extended hypotheses, serving as a lower bound limit for rescoring performance. 
\end{itemize}

As shown in Table \ref{table:1}, ProGRes outperforms both prompt-only and LLM rescoring approaches. We observe an improvement in 87.5\% of the experiments in the table, which consider various LLMs and ASRs.
Among the LLMs tested, GPT-4 achieves the highest performance, 
followed by GPT-3.5 and Llama-3. For instance, GPT-4 achieves a relative performance improvement over the baseline of 4.9\% for ASR$_1$ (with a WER of 42.94\% for the baseline and 40.84\% for ProGRes) and 24.7\% for ASR$_2$ (with a WER of 12.38\% for the baseline and 9.82\% for ProGRes).
This result suggests that larger improvements are achievable when the ASR hypotheses are only slightly inaccurate (ASR$_2$). Conversely, in mismatched training scenarios where ASR hypotheses contain numerous errors (ASR$_1$), LLMs struggle to identify suitable corrections, as highlighted by the minimal improvement in the prompt-only configuration for ASR$_1$ compared to the baseline (42.65\% vs 42.94\%). We also include oracle WERs, with a baseline oracle WER of 38.38\% for ASR$_1$ and 6.82\% for ASR$_2$, which decreases to a minimal oracle WER of 36.17\% for ASR$_1$ and 5.48\% for ASR$_2$ with GPT-4 prompting.

The results in Table \ref{table:1} also reveal that relying solely on LLM rescoring does not lead to substantial improvements in ASR accuracy, confirming the necessity of integrating both LLM and ASR scores. It is worth highlighting that, despite the considerable performance enhancement over the baseline, there remains a significant gap compared to the oracle. This result suggests that even after rescoring, the optimal choice is not consistently identified.

We also considered using forced alignment to generate the ASR score for the prompt addition. In our preliminary experiments, using a subset of 3064 audio files from the validation set, we saw a reduction in WER from 8.03\% with our existing approach to 7.75\% using forced alignment scores. Though this process adds an extra layer of computation, it is a more principled way of scoring prompt additions instead of using the same score as the ASR 1-best, especially for faulty additions.

\begin{figure*}[t!]
    \centering
    \begin{subfigure}[b]{0.495\textwidth}
        \centering
        \includegraphics[width=1.0\linewidth]{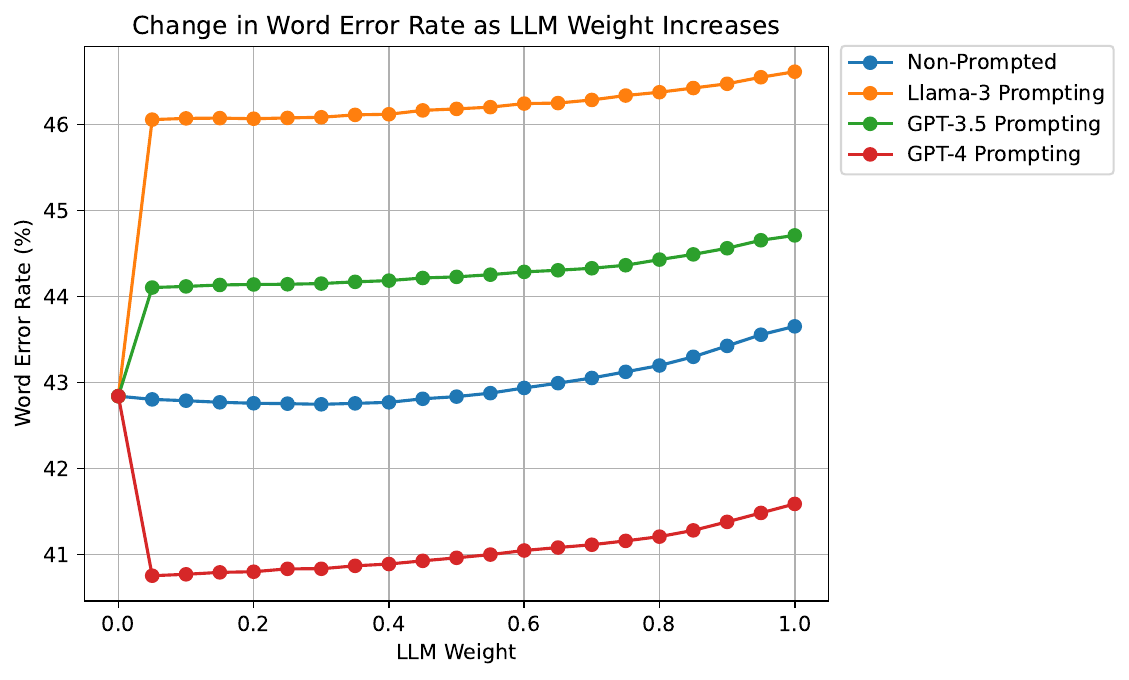}
        \label{fig:test1}
    \end{subfigure}
    \begin{subfigure}[b]{0.495\textwidth}
        \centering
        \includegraphics[width=1.0\linewidth]{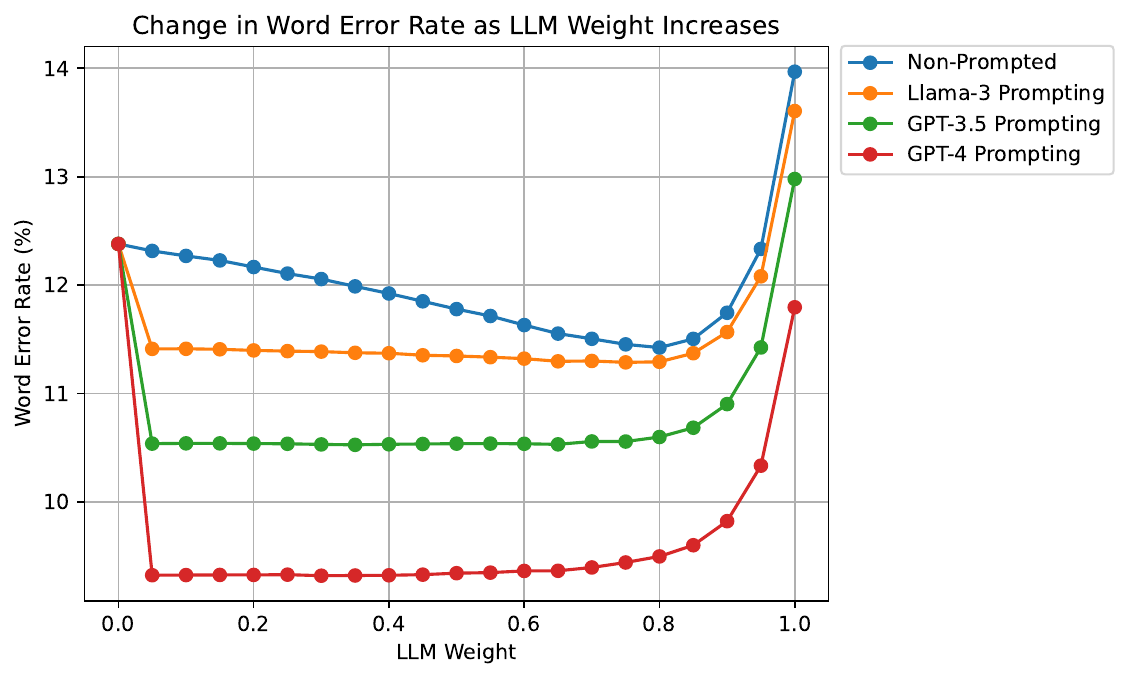}
        \label{fig:test2}
    \end{subfigure}
    \caption{WER results for different language model weights $\alpha$. The left panel shows results for ASR$_1$, and the right panel shows results for ASR$_2$. Non-prompted results simply refer to the original ASR $n$-best hypotheses.}
    \label{fig:wer_results}
    \vspace*{-2ex}
\end{figure*}

\vspace*{-2ex}
\subsection{Effect of LLM Weight}
\label{sec:int}
The weight assigned to the LLM scores during rescoring balances the contributions of ASR and LLM scores through linear interpolation, as discussed in Sec. \ref{sec:interpolation}.
The results in Table \ref{table:1} utilize the LLM weight optimized through a line search on the validation set.
Figure \ref{fig:wer_results} illustrates the impact of varying the LLM weight from 0 (ASR rescoring only) to 1 (LLM rescoring only) with increments of 0.05. 

Interestingly, the performance curves remain relatively flat across a broad range for both ASR$_1$ (Figure \ref{fig:wer_results}-left) and ASR$_2$ (Figure \ref{fig:wer_results}-right), as well as for different LLMs.
We observed that setting the LLM weight to a moderate value, neither too small (e.g., $\alpha \geq $0.05) nor too large (e.g., $\alpha \leq$0.6), provides a remarkable performance improvement compared to the non-rescored case ($\alpha=0$).

This evidence confirms the quality of the hypotheses generated by the LLM, as they frequently achieve the highest combined scores; as shown in Table \ref{table:1}, the addition of these LLM-generated hypotheses outperforms the original $n$-best with WERs from 13.61\% to 11.80\% compared to 13.97\%. These hypotheses are often linguistically plausible, and even a small boost can result in them achieving the best scores. However, the ProGRes system, through the combination of ASR and LLM scores, is still able to filter out noisy cases (e.g, when the LLM-generated hypotheses are nonsensical or incomplete) or output ASR hypotheses that are a better prediction than the prompt addition.
\vspace*{-2ex}
\subsection{Prompt Tuning}
We observed that the choice of prompt significantly influences the hypotheses generated by the LLM. Initially, we considered the following prompt:

\footnotesize
\begin{verbatim}
"Below are hypotheses for the transcription of a 
natural language sentence produced by an automatic 
speech recognizer. Please return only the sentence 
you believe is the most plausible transcription, 
surrounding the sentence with <>. Reminder that 
spelling of names may be different, as well as words
used, as the hypotheses are phonetically estimated,
so please return a sentence that is NOT in the
hypotheses. Please provide the most likely based on 
information you have about plausible sentences from
these n-best: [ASR n-best]"
\end{verbatim}
\normalsize

This prompt appears reasonable and apparently similar to those used by Ma et al. in prior unconstrained experiments (see Sec. \ref{sec:prompted}). However, we found it suboptimal in some scenarios. Specifically, the prompt performed well overall, but it was prone to cause hallucinations, in the form of unnecessary sentence extensions, repeated sentences, or unrelated additions.
After experimenting with the prompt proposed by Ma et al., we observed a lower rate of hallucinations and a closer alignment of generated hypotheses with the ASR outputs. This difference can be attributed to the varying degrees of constraint imposed by the prompts. The initial prompt allowed more freedom to the LM, thereby increasing the risk of hallucinations. This result highlights the crucial role of prompt engineering in the context of ASR $n$-best rescoring.
\vspace*{-2ex}
\subsection{Qualitative Example}
In this section, we present a qualitative example that illustrates the impact of LLM-generated hypotheses. When the best hypothesis differs from the ASR's top choice, the prompted LLM can exploit information from other hypotheses to generate an improved hypothesis. For instance, consider the following example:
\begin{itemize} 
\itemsep -0.5ex 
\item \textbf{ASR 1-BEST}: my knee wants to zoom 
\item \textbf{PROMPT ADDITION}: my niece wants tickets to zoo 
\item \textbf{GROUND TRUTH}: my niece wants tickets to the zoo 
\end{itemize}
Initially, the prompt addition may seem unlikely: how could the prompted model derive ``my niece" from ``my knee" and ``tickets to zoo" from ``to zoom"? However, upon examining the ASR $n$-best list for this specific example, the process becomes clear:
\begin{itemize} \itemsep -1ex 
\item \textit{my knee wants to zoom} 
\item \textit{my knee wants to zoe}
\item \textit{my knee's wants to zoom} 
\item \textit{my knee's wants to zoe} 
\item \textit{my knee's wants tickets to zoom} 
\item \textit{my knee's wants tickets to zoe} 
\item \textit{my knee's wants to zoom and} 
\end{itemize}
The prompted model analyzes these options and formulates a plausible hypothesis. It recognizes that ``tickets to zoom/zoe" can be interpreted as ``tickets to zoo", and it corrects ``knee's" to ``niece" based on grammatical context. This example demonstrates how the prompted model effectively utilizes the ASR $n$-best alternatives and its linguistic knowledge to propose a highly accurate option for rescoring purposes.
\vspace*{-2ex}
\subsection{Contamination}
One crucial factor to consider when evaluating LLMs on large, open-source datasets is the risk of contamination. Its performance may not accurately reflect real-world scenarios if the LLM model is trained on the considered validation and testing data.
For instance, like Ma et al., our study utilizes CommonVoice, a dataset comprised of substantial text transcriptions from Wikipedia, which is commonly incorporated into the training datasets of modern LLMs.
We regard the knowledge acquired from sources like Wikipedia and other web-based resources as part of the ``world knowledge" learned by the LLM. As shown in our results, this knowledge integration plays a crucial role in improving the ASR performance.

However, we perform a sanity check to assess the level of contamination in our prompt additions. Our primary concern is ensuring that prompt additions do not merely replicate accurate transcripts already available on Wikipedia. The most straightforward method for this analysis involves searching for named entities that are unrelated to common cultural knowledge. For instance, the model should correctly identify ``Elvis" when the sentence provides clues like ``the King of Rock and Roll". If the model suggests ``Jon" instead of ``John", and the transcription consistently uses ``Jon" without any supporting context in the hypotheses, it suggests potential contamination. This occurs when the model corrects a specific named entity from the data that the ASR does not include in its $n$-best results.

To analyze this aspect, we filtered instances where the prompt addition was selected as the top candidate and compare it with the ASR's 1-best output for the same audio sample and its transcription. We identified ``rare" named entities from the transcriptions that are absent in the ASR's $n$-best results and verified that these are also absent in the prompt additions.
 Below, we provide some examples where we checked contamination within data-specific named entities:

\small
\begin{itemize}[leftmargin=*]
  \item
      \textbf{ASR 1-BEST:} \underline{\textit{copper}} is married with 2 children.\\
      \textbf{PROMPT ADDITION:} \underline{\textit{copper}} is married with two children. \\
      \textbf{GROUND TRUTH:} \underline{\textit{popov}} is married with two children. 
  \item
        \textbf{ASR 1-BEST:}  \underline{\textit{mccleann}} a graduate of sequoia high school and the college of sanmato. \\
      \textbf{PROMPT ADDITION:}  \underline{\textit{mccleann}} a graduate of sequoia high school and the college of san mateo.  \\
      \textbf{GROUND TRUTH:}  \underline{\textit{mcclellan}} a graduate of sequoia high school and the college of san mateo. 
  \item
        \textbf{ASR 1-BEST:} his yogurt sister \underline{\textit{farland}} plagued valueball for the canadian national team. \\
      \textbf{PROMPT ADDITION:} his younger sister \underline{\textit{farland}} played volleyball for the canadian national team. \\
      \textbf{GROUND TRUTH:} his younger sister \underline{\textit{falin}} played volleyball for the canadian national team. 
  \item
      \textbf{ASR 1-BEST:} \underline{\textit{third janssen}} has developed what is called to intelligent design. \\
      \textbf{PROMPT ADDITION:} \underline{\textit{third janssen}} has developed what is called the intelligent design group. \\
        \textbf{GROUND TRUTH:} \underline{\textit{phillip johnson}} has developed what is called the intelligent design movement. 
\end{itemize}

\normalsize
In none of the cases above had the LLM generated the ground truth named entity. Based on our analysis, we found no evidence that data-specific knowledge beyond the hypotheses influenced the prompt additions.

\vspace*{-2ex}
\section{Conclusions}
This work introduces ProGRes, a novel method designed for improving ASR $n$-best rescoring by leveraging information from modern LLMs. Our approach extends the initial set of $n$-best hypotheses provided by the ASR with hypotheses generated by prompt-based LLMs. ProGRes then reranks this extended set of hypotheses using scores derived from both linguistic information (obtained via open-weight LLMs) and acoustic information (provided by the ASR).
We evaluated our approach using various LLMs and ASRs and demonstrated significant reductions in WER. Specifically, on the CommonVoice test set, we observed a WER reduction from 42.94\% to 40.84\% for the first ASR system and 12.38\% to 9.32\% for the second ASR.

One limitation of our approach is its computational complexity, primarily due to the resource-intensive nature of LLMs. Nevertheless, we believe that ProGRes can applied in domains where transcription accuracy is critical, such as medical dictations and meeting and lecture transcriptions involving technical or uncommon terminology.

In future research, we plan to explore new methods to reduce resource consumption while effectively taking advantage of the rich information contained within LLMs.
We also plan to explore fine-tuning generative models on the aforementioned domains, exploring more diverse ASRs and datasets, and continuing analysis with forced alignment.

{\bf Acknowledgments:}
We acknowledge the support of the Natural Sciences and Engineering Research Council of Canada (NSERC) and the Digital Research Alliance of Canada (alliancecan.ca). We thank OVHCloud for donating part of the GPU computing resources needed for this work.

\bibliographystyle{IEEEbib}
\bibliography{bibliography}

\begin{thebibliography}{10}

\bibitem{yu2018recent}
D.~Yu and J.~Li,
\newblock ``Recent progresses in deep learning based acoustic models,''
\newblock {\em IEEE/CAA Journal of Automatica Sinica}, vol. 5, no. 3, pp. 409--419, 2018.

\bibitem{li2014overview}
J.~Li, L.~Deng, Y.~Gong, and R.~Haeb-Umbach,
\newblock ``An overview of noise-robust automatic speech recognition,''
\newblock {\em IEEE/ACM Transactions on Audio, Speech, and Language Processing}, vol. 22, no. 4, pp. 745--777, 2014.

\bibitem{kinoshita2016summary}
K.~Kinoshita, M.~Delcroix, S.~Gannot, et~al.,
\newblock ``{A Summary of the REVERB Challenge: State-of-the-Art and Remaining Challenges in Reverberant Speech Processing Research},''
\newblock {\em EURASIP Journal on Advances in Signal Processing}, vol. 7, 2016.

\bibitem{noise_cont}
M.~Ravanelli and M.~Omologo,
\newblock ``{Contaminated Speech Training Methods for Robust DNN-HMM Distant Speech Recognition},''
\newblock in {\em Proceedings of Interspeech}, 2015.

\bibitem{reverb}
M.~Ravanelli and M.~Omologo,
\newblock ``{On the Selection of the Impulse Responses for Distant-Speech Recognition Based on Contaminated Speech Training},''
\newblock in {\em Proceedings of Interspeech}, 2014.

\bibitem{matassoni2014dirha}
M.~Matassoni, R.~Astudillo, A.~Katsamanis, and M.~Ravanelli,
\newblock ``{The DIRHA-GRID Corpus: Baseline and Tools for Multi-Room Distant Speech Recognition Using Distributed Microphones},''
\newblock in {\em Proceedings of Interspeech}, 2014.

\bibitem{ravanelli2017deep}
Mirco Ravanelli,
\newblock {\em Deep Learning for Distant Speech Recognition},
\newblock Ph.D. thesis, University of Trento, 2017.

\bibitem{cornell2023chime7}
S~Cornell et~al.,
\newblock ``The chime-7 dasr challenge: Distant meeting transcription with multiple devices in diverse scenarios,''
\newblock {\em arXiv preprint arXiv:2306.13734}, 2023.

\bibitem{info14020137}
P.~Gabler, B.~C. Geiger, B.~Schuppler, and R.~Kern,
\newblock ``{Reconsidering Read and Spontaneous Speech: Causal Perspectives on the Generation of Training Data for Automatic Speech Recognition},''
\newblock {\em Information}, vol. 14, no. 2, 2023.

\bibitem{mousavi2024are}
Seyed~Mahed Mousavi, Gabriel Roccabruna, Simone Alghisi, Massimo Rizzoli, Mirco Ravanelli, and Giuseppe Riccardi,
\newblock ``{Are LLMs Robust for Spoken Dialogues?},''
\newblock in {\em Proceedings of the International Workshop on Spoken Dialogue Systems Technology (IWSDS)}, 2024.

\bibitem{liu2011entity}
X.~Liu, M.~J. Gales, and P.~C. Woodland,
\newblock ``Using named entities in speech recognition,''
\newblock in {\em Proceedings of Interspeech}, 2011.

\bibitem{caubriere-etal-2020-named}
A.~Caubri{\`e}re et~al.,
\newblock ``Where are we in named entity recognition from speech?,''
\newblock in {\em Proceedings of LREC}, 2020.

\bibitem{mdhaffar2022end}
S.~Mdhaffar, J.~Duret, T.~Parcollet, and Y.~Estève,
\newblock ``{End-to-End Model for Named Entity Recognition from Speech Without Paired Training Data},''
\newblock in {\em Proceedings of Interspeech}, 2022.

\bibitem{asr-n-best-rescoring}
E.~Kang, C.~Van Gysel, and M.-H. Siu,
\newblock ``Transformer-based model for asr n-best rescoring and rewriting,''
\newblock in {\em Proceedings of Interspeech}, 2024.

\bibitem{languages7030236}
J.~Jansen~van Vüren and T.~Niesler,
\newblock ``Improving n-best rescoring in under-resourced code-switched speech recognition using pretraining and data augmentation,''
\newblock {\em Languages}, vol. 7, no. 3, 2022.

\bibitem{9747118}
L.~Xu, Y.~Gu, J.~Kolehmainen, H.~Khan, A.~Gandhe, A.~Rastrow, A.~Stolcke, and I.~Bulyko,
\newblock ``Rescorebert: Discriminative speech recognition rescoring with bert,''
\newblock in {\em Proceedings of ICASSP}, 2022.

\bibitem{udagawa2022effect}
T.~Udagawa, M.~Suzuki, G.~Kurata, N.~Itoh, and G.~Saon,
\newblock ``Effect and analysis of large-scale language model rescoring on competitive {ASR} systems,''
\newblock in {\em Proceedings of Interspeech}, 2022.

\bibitem{9679032}
Y.~Song, X.~Huang, X.~Zhao, D.~Jiang, and R.~Wong,
\newblock ``Multimodal n-best list rescoring with weakly supervised pre-training in hybrid speech recognition,''
\newblock in {\em Proceedings of ICDM}, 2021.

\bibitem{Xu2018APR}
H.~Xu, T.~Chen, D.~Gao, Y.~Wang, K.~Li, N.~Kumar Goel, Y.~Carmiel, D.~Povey, and S.~Khudanpur,
\newblock ``A pruned rnnlm lattice-rescoring algorithm for automatic speech recognition,''
\newblock {\em Proceedings of ICASSP}, 2018.

\bibitem{chung-etal-2012-lattice}
E.~Chung, H.~Jeon, J.~Park, and Y.~Lee,
\newblock ``Lattice rescoring for speech recognition using large scale distributed language models,''
\newblock in {\em Proceedings of {COLING} 2012}, 2012.

\bibitem{6707767}
Fl~Peng, Sl~Roy, B.~Shahshahani, and F.~Beaufays,
\newblock ``Search results based n-best hypothesis rescoring with maximum entropy classification,''
\newblock in {\em Proceedings of ASRU}, 2013.

\bibitem{DBLP:journals/corr/abs-1210-8440}
C.~Chelba, D.~Bikel, M.~Shugrina, P.~Nguyen, and S.~Kumar,
\newblock ``Large scale language modeling in automatic speech recognition,''
\newblock {\em arXiv preprint arXiv:1210.8440}, 2012.

\bibitem{Sak2010OntheflyLR}
H.~Sak, M.~Saraçlar, and Tunga G.,
\newblock ``{On-the-fly lattice rescoring for real-time automatic speech recognition},''
\newblock in {\em Proceedings of Interspeech}, 2010.

\bibitem{ma23}
R.~Ma, M.. Gales, K.~Knill, and M.~Qian,
\newblock ``N-best {T5}: Robust {ASR} error correction using multiple input hypotheses and constrained decoding space,''
\newblock in {\em Proceedings of Interspeech}, 2023.

\bibitem{ORTMANNS199743}
S.~Ortmanns, H.~Ney, and X.~Aubert,
\newblock ``A word graph algorithm for large vocabulary continuous speech recognition,''
\newblock {\em Computer Speech and Language}, vol. 11, no. 1, pp. 43--72, 1997.

\bibitem{hakkani2003}
D.~Hakkani-Tur and G.~Riccardi,
\newblock ``A general algorithm for word graph matrix decomposition,''
\newblock in {\em Proceedings of ICASSP}, 2003.

\bibitem{vaswani2023attention}
A.~Vaswani, N.~Shazeer, N.~Parmar, J.~Uszkoreit, L.~Jones, A.~N Gomez, L.~Kaiser, and I.~Polosukhin,
\newblock ``Attention is all you need,''
\newblock in {\em Proceesings of NeurIPS}, 2017.

\bibitem{ma2023generative}
R.~Ma, M.~Qian, P.~Manakul, M.~Gales, and K.~Knill,
\newblock ``Can generative large language models perform {ASR} error correction?,''
\newblock {\em arXiv preprint arXiv:2307.04172}, 2023.

\bibitem{fohr21}
D.~Fohr and I.~Illina,
\newblock ``Bert-based semantic model for rescoring n-best speech recognition list,''
\newblock in {\em Proceedings of Interspeech}, 2021.

\bibitem{namazifar2021correcting}
M.~Namazifar, J.~Malik, L.~Erran Li, G.~Tur, and D.~Hakkani-Tür,
\newblock ``Correcting automated and manual speech transcription errors using warped language models,''
\newblock {\em arXiv preprint arXiv:2103.14580}, 2021.

\bibitem{watanabe2018espnet}
S.~Watanabe et~al.,
\newblock ``{ESPnet}: End-to-end speech processing toolkit,''
\newblock in {\em Proceedings of Interspeech}, 2018.

\bibitem{hayashi2020espnet}
T.~Hayashi et~al.,
\newblock ``{Espnet-TTS}: Unified, reproducible, and integratable open source end-to-end text-to-speech toolkit,''
\newblock in {\em Proceedings of ICASSP}, 2020.

\bibitem{ravanelli2021speechbrain}
M.~Ravanelli et~al.,
\newblock ``Speechbrain: A general-purpose speech toolkit,''
\newblock {\em arXiv preprint arXiv:2106.04624}, 2021.

\bibitem{baevski2020wav2vec}
A.~Baevski, H.~Zhou, A.~Mohamed, and M.~Auli,
\newblock ``wav2vec 2.0: a framework for self-supervised learning of speech representations,''
\newblock in {\em Proceedings of NeurIPS}, 2020.

\bibitem{gulati2020conformer}
A.~Gulati et~al.,
\newblock ``Conformer: Convolution-augmented transformer for speech recognition,''
\newblock in {\em Proceedings of Interspeech}, 2020.

\bibitem{sennrich-etal-2016-neural}
R.~Sennrich, B.~Haddow, and A.~Birch,
\newblock ``{Neural Machine Translation of Rare Words with Subword Units},''
\newblock in {\em Proceedings of ACL}, 2016.

\bibitem{openai2024gpt4technicalreport}
OpenAI et~al.,
\newblock ``Gpt-4 technical report,'' 2024.

\bibitem{brown2020language}
T.~Brown et~al.,
\newblock ``{Language Models are Few-Shot Learners},''
\newblock in {\em Proceedings of NeurIPS}, 2020.

\bibitem{touvron2023llama}
Hugo Touvron et~al.,
\newblock ``{LLaMA:} open and efficient foundation language models,''
\newblock {\em arXiv preprint arXiv:2302.13971}, 2023.

\bibitem{commonvoice:2020}
R.~Ardila et~al.,
\newblock ``Common voice: A massively-multilingual speech corpus,''
\newblock in {\em Proceedings of LREC}, 2020.

\bibitem{torchaudio21}
J.~Hwang et~al.,
\newblock ``{TorchAudio 2.1: Advancing speech recognition, self-supervised learning, and audio processing components for PyTorch},''
\newblock in {\em Proceedings of ASRU}, 2023.

\bibitem{misra2022minicons}
Kanishka Misra,
\newblock ``minicons: Enabling flexible behavioral and representational analyses of transformer language models,''
\newblock {\em arXiv preprint arXiv:2203.13112}, 2022.

\end{thebibliography}

\end{document}